# Approximations in Bayesian Belief Universes for Knowledge-Based Systems

*Frank Jensen  &  Stig Kjær Andersen*
Institute of Electronic Systems
Aalborg University
Fr. Bajersvej 7, DK-9220 Aalborg Ø, Denmark

## Abstract

When expert systems based on causal probabilistic networks (CPNs) reach a certain size and complexity, the "combinatorial explosion monster" tends to be present. We propose an approximation scheme that identifies rarely occurring cases and excludes these from being processed as ordinary cases in a CPN-based expert system. Depending on the topology and the probability distributions of the CPN, the numbers (representing probabilities of state combinations) in the underlying numerical representation can become very small. Annihilating these numbers and utilizing the resulting sparseness through data structuring techniques often results in several orders of magnitude of improvement in the consumption of computer resources. Bounds on the errors introduced into a CPN-based expert system through approximations are established. Finally, reports on empirical studies of applying the approximation scheme to a real-world CPN are given.

**Keyword**s: Approximative reasoning, belief network, causal probabilistic network, expert system, knowledge-based system, influence diagram, junction tree, probability propagation, reasoning under uncertainty.

## 1  Introduction

Expert systems, using causal probabilistic networks (CPNs)[1] for knowledge representation, are reaching the state where it is feasible to handle domains modeled by large-scale networks (e.g., MUNIN [Andreassen *et al.*, 1987; Olesen *et al.*, 1989]). When building such large networks, it is (for reasons of practicality) often necessary to introduce approximations besides those inherent in the process of modeling a domain. Two main approaches have been investigated: focusing on the development of an approximative algorithm for propagation of information (e.g., [Henrion, 1989]), and focusing on approximations in the underlying network representation and then using an exact inference algorithm.

The objective of this paper is to present an approximation scheme that takes the latter approach. The scheme is tailored to the Bayesian belief universe approach [Jensen *et al.*, 1989] as used in HUGIN [Andersen *et al.*, 1989]. The method operates by approximations in the quantitative part of the underlying representation, whereas the qualitative structure remains unchanged. Within this framework, we can assess the accuracy of the approximated probabilities, which is not possible with heuristic methods. Application of the method often results in a substantial decrease in the usage of computer resources; the amount of decrease depends on domain characteristics, such as network topology and probability distributions.

It is known that, in general, probabilistic inference in CPNs is $\mathcal{NP}$-hard [Cooper, 1987], and exact calculations will eventually become intractable. This fact emphasizes the importance of approximative methods.

A domain model in the causal probabilistic network approach consists of a graph with nodes representing the domain variables and the (directed) arcs representing the causal relations between the domain variables. Conditional probabilities are used to describe the dependency of domain variables given their immediate predecessors (parents). Different inference methods have been developed to propagate information in such a network: If the topology is simple (singly connected) [Pearl, 1986], propagation can be done directly in the CPN; otherwise, a secondary structure for topologies, including nondirected loops [Lauritzen and Spiegelhalter, 1988; Jensen *et al.*, 1989; Shafer and Shenoy, 1988], can be used. Alternatively, for the latter kind of topologies, the inference could also take place in a set of conditioned networks [Suermondt and Cooper, 1988] or through manipulation of the network with an arc-reversing technique [Shachter, 1988].

The method of Bayesian belief universes splits the inference task into two phases: a compilation phase and a run-time phase. The proposed approxima-

---

[1] Synonyms: belief networks, causal networks, and probabilistic influence diagrams.

tion scheme adds another phase to this task: The *approximation and compression* phase. The phases are thus

- *The compilation phase*: Based on the CPN domain model, a secondary structure is constructed—a so-called *junction tree* of belief universes.
- *The approximation and compression phase*: Small numbers, representing the probabilities of very rare cases, are annihilated (set to zero), thereby effectively eliminating these cases from the domain model. Through use of data structuring techniques for sparse tables, the underlying numerical tables (the *belief tables*) of the junction tree are compressed.
- *The run-time phase*: The actual inference takes place in the junction tree, using the modified belief tables.

In Section 2, we review the basic belief universe concepts essential for the proposed approximation scheme. Section 3 describes how to perform the approximation and establishes some worst-case error bounds on probabilities obtained from the approximated junction tree. Finally, Section 4 reports empirical results we obtained by applying the proposed approximation scheme to a real-world CPN—namely, one of the networks of the MUNIN knowledge base.

## 2 Belief Universes

This section reviews some of the basic concepts of the belief universe approach.

The domain represented by the CPN is divided into a set of subdomains called *belief universes*. A belief universe $U$ consists of two parts: a *set of nodes*[2] and a *belief table*, which contains an assessment of the joint probabilities for the state space of $U$ (i.e., the Cartesian product of the state sets for the nodes of $U$).

The construction of a system of belief universes, equivalent to the original CPN domain model, consists of the following steps:

- *Form the moral graph*: For each node in the network, add links between all of its parents that are not already linked. Drop the directions.
- *Triangulate*[3] *the moral graph*: Add links to the moral graph until a triangulated graph is obtained.
- *Form the system of belief universes*: The node sets are the cliques[4] of the triangulated graph. The initial belief tables are calculated as appropriate products of the conditional probability tables [Lauritzen and Spiegelhalter, 1988; Jensen *et al.*, 1989].
- *Organize the system as a junction tree*: Links between belief universes are introduced, such that a tree with the following property results: For each pair $(U, V)$ of belief universes, each belief universe on the unique path between $U$ and $V$ contains the nodes $U \cap V$. As shown in [Jensen, 1988], a junction tree can be constructed by a maximal spanning-tree algorithm.

All steps except the second are deterministic: There is only one moral graph, and the set of cliques of a triangulated graph is unique. There may be several junction trees, but the differences among them are minor (the major cost of a junction tree is the representation of the belief tables for the belief universes). The second step is important: A good triangulation can save substantial space and time [Kjærulff, 1990].

Let $U$ be a belief universe with belief table $B$, and let $S \subset U$. We can obtain the joint probabilities for $S$ from $B$ by summing up all beliefs in $B$ for $S$. This operation is called *marginalization*. In particular, the belief in a single node can be obtained by marginalization of the belief table of any belief universe containing it.

Let $U$ be a belief universe, and let $V \subseteq U$. A *finding* on $V$ is a subset of the state space of $V$.[5] The finding is *entered* into $U$[6] through annihilation of the elements in the belief table of $U$ corresponding to state combinations *not* in $V$.

A set of one or more findings is called a *case*.

A junction tree is said to be *consistent* if marginalization of two distinct belief universes $U$ and $U'$ with respect to some set of nodes $V$ (contained in both $U$ and $U'$) yield "identical" (i.e., proportional) results. This property is (re)established through the *global propagation* operation. This operation refers to a *local* propagation method for transmitting evidence between neighbors in a junction tree.

*Absorption* is the local propagation method: If we have entered evidence into a belief universe $V$, then an adjacent belief universe $U$ *absorbs* from $V$ through the following steps:

1. Calculate the belief table for $U \cap V$ by marginalization of the belief table of $U$.

2. Calculate the belief table for the same intersection by marginalization of the belief table of $V$.

3. Multiply the belief table of $U$ by the ratio of the table achieved by Step 1 and the table achieved by Step 2.

---

[2] We shall use $U$ to denote both the belief universe itself and its set of nodes.

[3] A graph is *triangulated* if every cycle of length greater than three has a chord.

[4] A *clique* is a maximal set of nodes, all of which are pairwise linked.

[5] Typically, a finding is a statement that a node is known to be in a particular state.

[6] We shall also use the phrase "*evidence* is entered into $U$."

When absorbing from several neighbors simultaneously, these steps must proceed in "parallel" (implying use of the *same* version of the belief table of $U$ in Step 1).

Global propagation is described in terms of two operations: *CollectEvidence* and *DistributeEvidence*. *CollectEvidence* is used when evidence from the entire system must be propagated to a single belief universe $U$: $U$ asks neighbors to *CollectEvidence*; when they are done, $U$ absorbs from them. *DistributeEvidence* is used when evidence from a single belief universe $U$ must propagate to the entire system: $U$ asks each neighbor to absorb from $U$ and then *DistributeEvidence* to its other neighbors.

A global propagation operation consists of *CollectEvidence* operation followed by a *DistributeEvidence* operation initiated from an arbitrary belief universe.

*CollectEvidence* has an important property. Assume that we have a consistent and normalized junction tree, and that we enter evidence into some of the belief universes of the junction tree. If we invoke *CollectEvidence* from some belief universe $U$, then the normalizing constant for the belief table of $U$, after *CollectEvidence* has terminated, is equal to the (prior) probability of the evidence.

## 3 The Approximation Scheme

As described in the previous section, the numbers in the belief tables of the belief universes represent probabilities in joint probability distributions. One might expect that excluding the smallest numbers (representing rare state combinations) will lead to substantial improvements in the requirements of computer resources. In this section, we shall investigate some properties of such a scheme.

Assuming we have a consistent junction tree, an *approximation* is performed in the following way:

1. For each belief universe in the junction tree, we select some elements of its belief table and annihilate those; the rest are left unchanged.

2. The junction tree is made consistent again by a global propagation.

3. [Optional] The belief tables of the belief universes are compressed in order to take advantage of the introduced zeros. (This step will not be described here; see [Jensen and Andersen, 1990] for details.)

### How Do We Select the Numbers to Be Annihilated?

As previously mentioned, we are interested in the small numbers. A simple way to do the selection is to use a threshold value to separate the numbers to be annihilated from the numbers to be kept. However, we cannot choose a global threshold value, as the size of tables and their distribution of numbers may vary substantially. So instead we shall use a local threshold value for each table.

We observe that, annihilating an element of a belief table, corresponds to entering a finding that says that the state combinations, corresponding to this element, are "impossible" (or are considered uninteresting). Moreover, the sum of the annihilated elements in a given belief table is the *probability* of all the state combinations (the finding) corresponding to those elements. This probability is a measure of the (local) error, we commit. We can control this error by choosing a suitable threshold value.

Suppose we want to retain $1 - \varepsilon$ of the probability mass of each belief table. Then, a simple method is to compute a threshold value $\delta$ by repeatedly halving $\delta$ (using $\varepsilon$ as the initial value for $\delta$) until the sum of the elements less than $\delta$ is no greater than $\varepsilon$;[7] these elements will be annihilated (we believe that either all or no elements with the same value in a given table should be eliminated). A more costly method is to sort the elements of the table and to repeat annihilating the smallest number(s) as long as the sum of the annihilated numbers does not exceed $\varepsilon$.

The global error $e$ (the total amount of probability mass removed) is computed as $e = 1 - \mu$, where $\mu$ is the normalization constant found during the global propagation step of the approximation algorithm.

Given an arbitrary case, we can determine if it is one of the cases that have been *completely* excluded from consideration by detecting a zero normalization constant. The probability of such a case occurring (assuming the assessed conditional probabilities are correct) is $e$.

For each remaining case, some of the state combinations supporting the case may have been eliminated. The accumulated probability for those state combinations determines the error on the posterior probabilities as shown in the following.

### How Good Is the Approximation?

Assume that we have approximated the belief universes and have propagated the approximations throughout the junction tree. We now have a consistent junction tree.

Let $A$ denote the approximation performed, and let $F$ denote a set of findings to be entered into the (consistent) approximated junction tree. Entering such a set of findings is a common operation when using the junction tree (or rather the underlying CPN) as an expert system. After $F$ has been entered, and the junction tree has been made consistent by propagation, we want to query the system for probabilities of the form $P(H|F)$, where $H$ is some hypothesis.[8] However, the probabil-

---

[7] This method is used in Hugin [Andersen *et al.*, 1989].

[8] In a real application, the CPN might model the relationships between some diseases and the associated symptoms; $F$ then would be the set of symptoms found, $H$ typically would be of the form "the patient has disease $X$," and $P(H|F)$ would denote the probability that

ity $P(H|F)$ is not available; instead, we get the probability $P(H|F, A)$ (that is, the probability for $H$ given the findings $F$ and the approximation $A$).

We therefore want to find an upper bound on $|P(H|F) - P(H|F, A)|$:

$$|P(H|F) - P(H|F, A)|$$
$$= |P(H|F, A)P(A|F)$$
$$\quad + P(H|F, \overline{A})P(\overline{A}|F) - P(H|F, A)|$$
$$= |P(H|F, A)[P(A|F) - 1] + P(H|F, \overline{A})P(\overline{A}|F)|$$
$$= P(\overline{A}|F)|P(H|F, \overline{A}) - P(H|F, A)|$$
$$\leq P(\overline{A}|F)$$

The quantity $P(\overline{A}|F)$ can be rewritten as

$$\frac{P(F \cap \overline{A})}{P(F \cap \overline{A}) + P(F \cap A)}$$
$$= \frac{P(F \cap \overline{A})}{P(F \cap \overline{A}) + P(F|A)P(A)} \leq \frac{e}{e + \mu(1 - e)}$$

where $e = P(\overline{A})$ and $\mu = P(F|A)$. These quantities are known: $e$ is the approximation error found at approximation time, and $\mu$ is the normalization constant found during propagation of $F$. Unfortunately, $\mu$ is almost always small ($\ll e$), so this upper bound is not a good indicator of the approximation error.

In practice, however, $F$ is almost always of the form $f_1 \cap \ldots \cap f_n$, where $f_i$ ($1 \leq i \leq n$) states that "node $X_i$ is in state $y_i$." Thus

$$P(F \cap \overline{A}) \leq \min\{P(f_1 \cap \overline{A}), \ldots, P(f_n \cap \overline{A})\}$$

We can compute these quantities for all combinations of nodes and states at approximation time (the space required to store these quantities is small).

Although this gives us a better upper bound for the approximation error, it is, however, strictly a worst-case bound, and we may have to rely on empirical studies to determine the *actual* errors. In the next section, we shall investigate this issue for a real application.

## 4 An Application

We shall use a network from the MUNIN knowledge base to study the effect of the proposed approximation scheme on a real-world CPN.

The domain of MUNIN is electromyography, a technique for diagnosing peripheral muscle and nerve disorders. We have chosen a network describing disorders in the median nerve.[9] On the basis of four electromyographic findings, this model is capable of diagnosing three local nerve lesions and one diffuse disorder in the median nerve in the arm. The CPN contains 57 nodes; the disease nodes each have between three and five states, and the finding nodes have from 15 to 21 states.

The specification of the conditional probability tables requires 8126 numbers, of which 67.1 percent are assessed as zeros; however, most of these numbers have been generated by local models from a much smaller set of parameters, which has been assessed by domain experts [Andreassen *et al.*, 1987].

An explanation of the domain concepts, as well as a description of the medical performance, can be found in [Andreassen *et al.*, 1989; Olesen *et al.*, 1989].

### 4.1 Junction Trees

Based on different triangulations of the median-nerve CPN, we have created four junction trees, yielding different starting points for approximation. We have used a maximum-cardinality search [Tarjan and Yannakakis, 1984] and two heuristic search strategies that minimize the clique cardinality (the min-size heuristic) and the size of the state space of the nodes in the cliques (the min-weight heuristic), respectively; see [Kjærulff, 1990] for details.

| Clique Size | Triangulation Method | | | |
|---|---|---|---|---|
| | Max-Card 1 | Max-Card 2 | Min-Size | Min-Weight |
| | Number of Cliques | | | |
| 14 | 1 | - | - | - |
| 13 | 2 | - | - | - |
| 10 | 1 | - | - | - |
| 9 | 1 | - | - | - |
| 8 | 4 | 6 | 3 | 3 |
| 7 | 4 | 7 | 2 | 2 |
| 6 | 2 | 4 | 5 | 4 |
| 5 | 9 | 2 | 7 | 9 |
| Total State-space ($10^6$) | 4849 | 10.7 | 1.6 | 1.6 |
| Zeros (Percent) | - | 93 | 71 | 77 |
| Max State-space ($10^6$) | - | 4.0 | 0.45 | 0.54 |

Table 1: Statistics of junction trees for the median-nerve knowledge base generated from different triangulations.

Table 1 summarizes key parameters of junction trees, based on different triangulations. We have obtained two maximal-cardinality searches using differ-

---

the patient has disease $X$ given that he/she exhibits the symptoms $F$.

[9]It is our impression that this network is a "typical" network, in the sense that the benefits of approximation are neither negligible nor excessively large

ent starting nodes. However (for obvious reasons), we consider only the second one, referred to as "max-card," in the following subsections. The data in Table 1 apply to the initial consistent (i.e., after initialization) junction trees before any approximation or compression has been done.

## 4.2 Effect on Resources

We shall focus on two aspects of resources: (1) the propagation time needed to make the junction tree consistent after a set of findings has been entered, and (2) the storage space needed to represent the knowledge base in a suitable compact form (see [Jensen and Andersen, 1990] for details).

The global error $e$, defined in Section 3, is used to characterize the approximation; we shall use the term *total removed probability mass* to refer to this value.

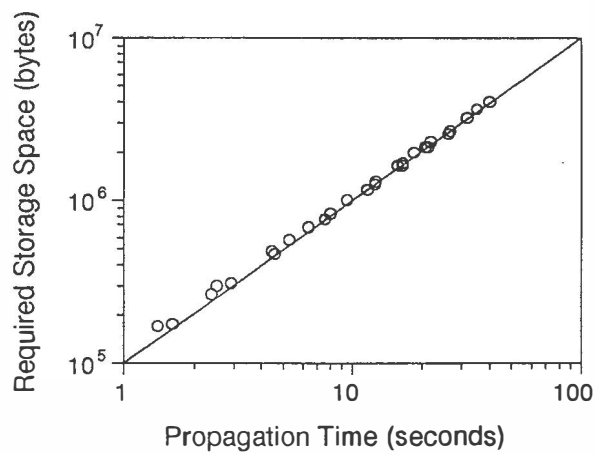

Figure 2: The relation between required storage space and propagation time for the median-nerve knowledge base for various approximations. The line corresponds to a linear relationship between propagation time and storage space.

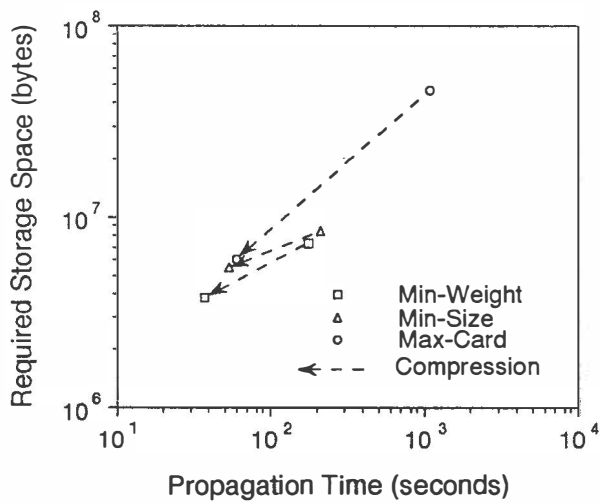

Figure 1: The effect of compression on required storage space and propagation time for the median-nerve knowledge base.

The time and space measurements reported are for an implementation of HUGIN [Andersen *et al.*, 1989] in C for a Sun 3 workstation; however, we are only interested in relative improvements, so the space and (in particular) time units should be regarded as arbitrary.

Figure 1 illustrates the effect of the initial compression on required storage space and propagation time for three different junction trees. As expected, the gain varies according to the different ratio of zeros in the junction trees (see Table 1).

Figure 2 shows the relation between propagation time and storage space needed for the three different triangulation methods at different approximations. The total removed probability mass ($e$) varied between 0.001 and 1 percent. At each data point, the corresponding approximated and compressed runtime system was created, and the time and space

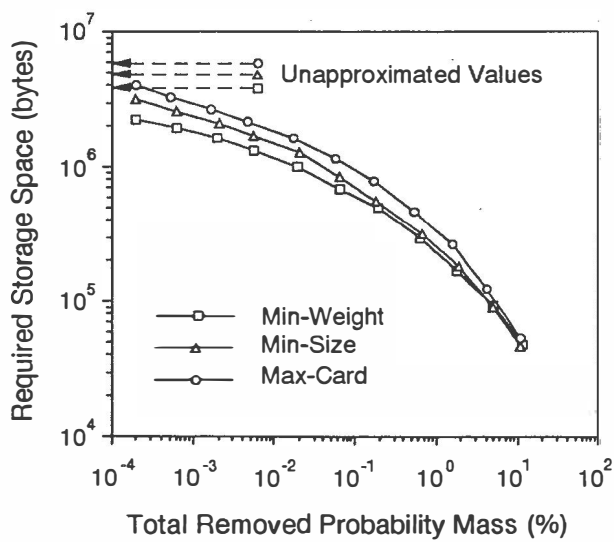

Figure 3: The space requirement as a function of the probability mass removed for different junction trees. The arrows indicate the storage requirements for unapproximated but compressed junction trees

characteristics were measured. We observe a linear relationship between propagation time and storage space needed; thus, we characterize resource requirements in term of storage space only.

The resource requirements for approximated junction trees as a function of the total removed probability mass is the subject of Figure 3. Each data point in this figure corresponds to a data point in Figure 2, except for points corresponding to $e > 2$ percent. The values corresponding to no approximation for the compressed junction trees are also indicated.

We observe that, for $e$ less than ~0.1 percent, the approximation is equally efficient for the three junction trees. For each junction tree, $e = 0.25$ percent yields about one order of magnitude in reduction of the required space. However, for a sufficiently large value of $e$, the differences between the junction trees disappear.

Table 2 shows the effect of the method applied to the different junction trees at $e = 0.1$ percent.

|  | Triangulation Method | | |
| --- | --- | --- | --- |
|  | Max-Card | Min-Size | Min-Weight |
| Space | | | |
| Initial (Mbytes) | 46 | 8.5 | 7.2 |
| Approx. (Mbytes) | 0.95 | 0.71 | 0.60 |
| Reduction | 0.989 | 0.916 | 0.916 |
| Time | | | |
| Initial (seconds) | 1100 | 213 | 175 |
| Approx. (seconds) | 9.5 | 7.1 | 6.0 |
| Reduction | 0.991 | 0.967 | 0.966 |

Table 2: The effect of approximation and compression on junction trees generated from the median-nerve CPN.

### 4.3 Effect on the Quality

Whenever we commit ourselves to making an approximation, we want to know the risk that we will make serious errors. Unfortunately, the basis on which we calculate the theoretical worst-case error bounds might be too coarse, and it is highly unlikely that the worst-case situation will appear in a real application. If we had some method that could warn us when the situation was questionable, we might take the risk and make approximations beyond the magnitude imposed by a given worst-case error bound. We shall use our median-nerve knowledge base, and shall make a diagnosis on the basis of a set of findings, thus showing how our theoretical estimate on upper bounds on errors compares to practical values.

Figure 4 displays the results of entering a typical case into various approximated junction trees. The probability of the case is $4.1 \times 10^{-4}$.

The observed error in the beliefs caused by the approximation is shown as a function of the total removed probability mass ($e$). The figure shows observed errors in the beliefs of states representing exact beliefs between 0.9189 and 0.0005. The worst-case error bound (Section 3) for each approximation and case also has been computed. We observe that the difference between the worst-case bound and the worst measured absolute error is about three orders of magnitude for $e \leq 0.1$ percent.

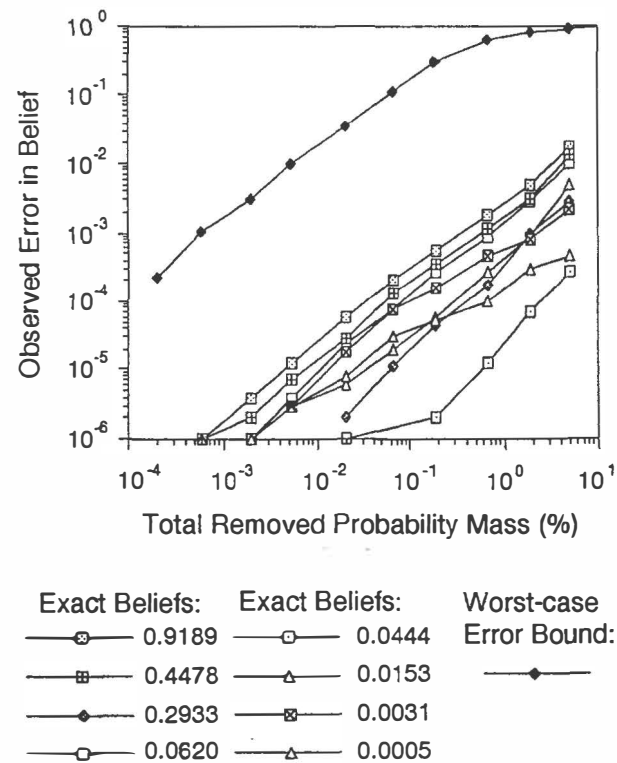

Figure 4: The errors observed in the beliefs for various states of a local nerve lesion given a standard case. The probability for this case is $4.1 \times 10^{-4}$.

Figure 5 shows triples of the worst-case bound (filled square), maximal observed error (diamond), and average observed error (open square) for 18 different randomly generated cases as a function of the case-specific normalizing constant, $\mu_{case}$. The approximation used corresponds to a decrease in resource requirements by a factor of four relative to an unapproximated but compressed junction tree.

Figure 5 shows that the observed errors on computed beliefs for the displayed cases are much smaller than that predicted by the worst-case error bound derived in Section 3. This difference shows that it is very unlikely, by picking a randomly generated case with a given $\mu_{case}$, to get the worst-case configuration. In the present CPN, the ratio between the worst-case bound and the maximal observed error is three orders of magnitude for $\mu_{\text{case}} \geq 10^{-6}$. Decreasing the normalizing constant ($\mu_{\text{case}}$) implies increasing the error in beliefs for the specific case, as well as for the worst-case error. When $\mu_{\text{case}}$ approaches zero, the error in beliefs approaches one, corresponding to excluding the case from the domain model.

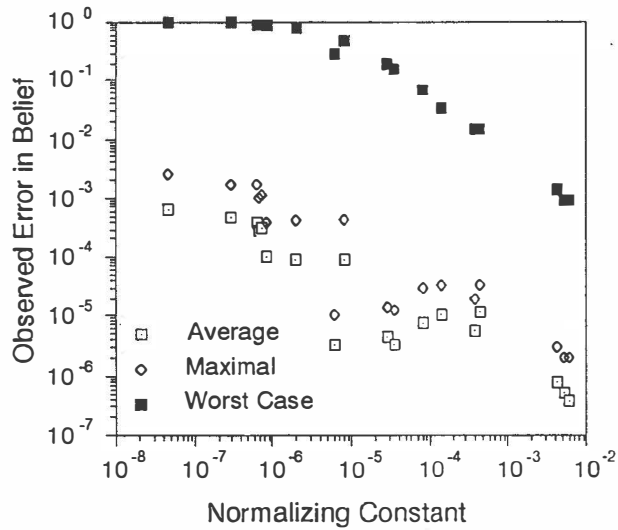

Figure 5: For $e = 2 \times 10^{-4}$, triples of worst-case error, maximal observed error, and average observed error in the beliefs of the states of the disorder nodes used for the case in Figure 4 are shown for 18 different cases.

These empirical studies show, that if we have a specific hypothesis in mind (for example the diagnosis of a local nerve lesion at the wrist) and a set of test cases which provides us with a span of $\mu_{\text{case}}$, we can get empirical values for the actual expected error in a specific case, given $\mu_{\text{case}}$.

Given a specific approximation $e$, we would have the following situations: If we insert a set of findings, and the theoretical worst-case error bound are below an accepted level, we can use the approximated junction tree. If we insert a set of findings which already has been taken out of the domain model by "zeroing out," the violation on the model will be recognized by a zero normalizing constant, and we have to use a less approximated junction tree. If we insert a set of findings yielding an unacceptably high worst-case error, we have to rely on empirical studies, such as those above, to estimate the error based on $\mu_{\text{case}}$, and on basis of this, decide whether to fall back on a less approximated junction tree or accept the risk of committing an error. This approach allow us to obtain a graceful degradation of the quality of diagnoses as the limit of the approximation is reached.

For the median-nerve knowledge base and the focus on the hypothesis of a lesion at the wrist, a demand of 0.01 as the upper limit of error in a state, would allow us set the alert threshold as low as $\mu_{\text{case}} = 10^{-7}$ for $e = 2 \times 10^{-4}$.

## 5 Conclusion

We have presented a scheme for approximation in the numerical part of a CPN-based expert system. Our approach eliminates the (small) numbers representing probabilities of rare combinations of findings, thereby preventing these findings from being treated as ordinary findings in the expert system. The approximation has two effects: (1) we may gain several orders of magnitude in improvement of resource usage, and (2) we may lose some accuracy in the computed beliefs. However, we can estimate case-specific upper bounds for the errors made on the computed beliefs, although these bounds may be too pessimistic, as the studies reported in Section 4 show.

If the case has been completely excluded by the approximation process, we will detect it by finding a zero normalizing constant during propagation; if the case is one of the common cases, we know that the computed beliefs can be trusted to a large degree. The problematic cases are the ones that have a nonzero probability outside the "trusted range" of probabilities (remember that the probability of a case is equal to the normalization constant found during propagation). We suggest that, when a problematic case occur, we should reenter the case into a less approximated (maybe even a nonapproximated) junction tree; however, this solution should rarely be necessary.

It would be nice to find an upper bound on the error of beliefs that is better (and still easily computable) than is the one presented in Section 3. Calculation of this bound involves the errors made on *individual* findings. We might be able to do better if we considered two or more findings simultaneously; however, a straightforward approach would require $O(s^n)$ space, where $s$ is the total number of states in the nodes, and $n$ is the number of findings considered.

There might be a clever technique to avoid considering all these combinations of findings and at the same time to provide a better error bound. We shall leave this topic for future research.

## 6 Acknowledgements

We thank Steffen L. Lauritzen, Kristian G. Olesen, and Finn V. Jensen for valuable comments, suggestions, and inspiring discussions on the subject of this paper.

We are grateful for the inspiring environment provided by the Medical Computer Science Group at Stanford University to one of us (SKA) from August, 1989, through June, 1990. Computer support was partly provided by the SUMEX-AIM resource, under NIH grant LM05208.

We also thank Lyn Dupré, Stanford University, for the many improvements of the prose she contributed to this paper.

## References

[Andersen et al., 1989] S. K. Andersen, K. G. Olesen, F. V. Jensen, and F. Jensen. "HUGIN—a shell for building Bayesian belief universes for expert systems." In *Proceedings of the Eleventh International Joint Conference on Artificial Intelligence*, pages 1080–1085, Detroit, Michigan, August 1989.

[Andreassen et al., 1987] S. Andreassen, M. Woldbye, B. Falck, and S. K. Andersen. "MUNIN—a causal probabilistic network for interpretation of electromyographic findings." In *Proceedings of the Tenth International Joint Conference on Artificial Intelligence*, pages 366–372, Milan, Italy, August 1987.

[Andreassen et al., 1989] S. Andreassen, F. V. Jensen, S. K. Andersen, B. Falck, U. Kjærulff, M. Woldbye, A. R. Sørensen, A. Rosenfalck, and F. Jensen. "MUNIN—an expert EMG assistant." In *Computer-Aided Electromyography and Expert Systems*, J. E. Desmedt (editor), Elsevier Science Publishers, Amsterdam, The Netherlands, 1989.

[Cooper, 1987] G. F. Cooper. "Probabilistic inference using belief networks is $\mathcal{NP}$-hard." Research Report KLS-87-27, Medical Computer Science Group, Stanford University, Stanford, California, 1987.

[Henrion, 1989] M. Henrion. "Towards efficient inference in multiply connected belief networks." In *Influence Diagrams, Belief Nets and Decision Analysis*, R. M. Oliver and J. Q. Smith (editors), John Wiley & Sons, Chichester, 1989.

[Jensen, 1988] F. V. Jensen. "Junction trees and decomposable hypergraphs." Judex Research Report, Judex Datasystemer A/S, Aalborg, Denmark, 1988.

[Jensen and Andersen, 1990] F. Jensen and S. K. Andersen. "Compact and efficient representations of belief tables in HUGIN." Research Report, Institute of Electronic Systems, Aalborg University, Aalborg, Denmark. Manuscript in preparation.

[Jensen et al., 1988] F. V. Jensen, K. G. Olesen, and S. K. Andersen. "An algebra of Bayesian belief universes for knowledge based systems." To appear in *Networks*.

[Jensen et al., 1989] F. V. Jensen, S. L. Lauritzen, and K. G. Olesen. "Bayesian updating in causal probabilistic networks by local computations." To appear in *Computational Statistics Quarterly*.

[Kjærulff, 1990] U. Kjærulff. "Triangulation of graphs — algorithms giving small total state space." Research Report R-90-09, Institute of Electronic Systems, Aalborg University, Aalborg, Denmark, 1990.

[Lauritzen and Spiegelhalter, 1988] S. L. Lauritzen and D. J. Spiegelhalter. "Local computations with probabilities on graphical structures and their application to expert systems." *Journal of the Royal Statistical Society, Series B (Methodological)*, 50(2):157–224, 1988.

[Olesen et al., 1989] K. G. Olesen, U. Kjærulff, F. Jensen, F. V. Jensen, B. Falck, S. Andreassen, and S. K. Andersen. "A MUNIN network for the median nerve—a case study on loops." *Applied Artificial Intelligence, Special Issue on Causal Modeling*, 3(2–3):385–403, 1989.

[Pearl, 1986] J. Pearl. "Fusion, propagation, and structuring in belief networks." *Artificial Intelligence*, 29(3):241–288, 1986.

[Shachter, 1986] R. D. Shachter. "Evaluating influence diagrams." *Operations Research*, 34(6): 871–882, 1986.

[Shachter, 1988] R. D. Shachter. "Probabilistic inference and influence diagrams." *Operations Research*, 36(4):589–605, 1988.

[Shafer and Shenoy, 1988] G. Shafer and P. Shenoy. "Bayesian and belief-function propagation." Working Paper 121, School of Business, University of Kansas, Lawrence, Kansas, 1988.

[Suermondt and Cooper, 1988] H. J. Suermondt and G. F. Cooper. "Updating probabilities in multiply connected networks." In *Proceedings of the Fourth Workshop on Uncertainty in Artificial Intelligence*, pages 335–343, Minneapolis, August 1988.

[Tarjan and Yannakakis, 1984] R. E. Tarjan and M. Yannakakis. "Simple linear-time algorithms to test chordality of graphs, test acyclicity of hypergraphs, and selectively reduce acyclic hypergraphs." *SIAM Journal on Computing*, 13(3): 566–579, 1984.

[Yannakakis, 1981] M. Yannakakis. "Computing the minimum fill-in is $\mathcal{NP}$-complete." *SIAM Journal on Algebraic and Discrete Methods*, 2(1):77–79, 1981.